\documentclass[letterpaper, 10 pt, conference]{ieeeconf}  

\IEEEoverridecommandlockouts                              

\usepackage{bm}
\usepackage{amsmath}
\usepackage{graphicx}
\usepackage{subcaption}
\usepackage{makecell}    
\usepackage{booktabs}
\usepackage[colorlinks, linkcolor=blue, urlcolor=blue]{hyperref}
\usepackage[font=small, labelfont=bf]{caption}

\title{\LARGE \bf
MRS-CWC: A Weakly Constrained Multi-Robot System with Controllable Constraint Stiffness for Mobility and Navigation in Unknown 3D Rough Environments}

\author{Runze Xiao$^{1}$, Yongdong Wang$^{1}$, Yusuke Tsunoda$^{2}$, Koichi Osuka$^{3}$, Hajime Asama$^{4}$
\thanks{$^{1}$Xiao and Wang are with the Department of Precision Engineering, School of Engineering, The University of Tokyo, Tokyo, Japan
        {\tt\small xiao@i-con.t.u-tokyo.ac.jp}}%
\thanks{$^{2}$Tsunoda Assistant Professor is with the Department of Mechanical Engineering, University of Hyogo, Hyogo, Japan}%
\thanks{$^{3}$Osuka Professor is with the Department of Mechanical Engineering, Osaka University, Osaka, Japan}%
\thanks{$^{4}$Asama Project Professor (Professor Emeritus), Tokyo College, Institute for Advanced Studies, The University of Tokyo, Tokyo, Japan}%
\thanks{Project Webpage:{\scriptsize \url{https://wyd0817.github.io/project-mrs-cwc/}}}
}

\begin{document}

\maketitle
\thispagestyle{empty}
\pagestyle{empty}

\begin{abstract}
Navigating unknown three-dimensional (3D) rugged environments is challenging for multi-robot systems. Traditional discrete systems struggle with rough terrain due to limited individual mobility, while modular systems—where rigid, controllable constraints link robot units—improve traversal but suffer from high control complexity and reduced flexibility. To address these limitations, we propose the Multi-Robot System with Controllable Weak Constraints (MRS-CWC), where robot units are connected by constraints with dynamically adjustable stiffness. This adaptive mechanism softens or stiffens in real time during environmental interactions, ensuring a balance between flexibility and mobility. We formulate the system’s dynamics and control model and evaluate MRS-CWC against six baseline methods and an ablation variant in a benchmark dataset with 100 different simulation terrains. Results show that MRS-CWC achieves the highest navigation completion rate and ranks second in success rate, efficiency, and energy cost in the highly rugged terrain group, outperforming all baseline methods without relying on environmental modeling, path planning, or complex control. Even where MRS-CWC ranks second, its performance is only slightly behind a more complex ablation variant with environmental modeling and path planning. Finally, we develop a physical prototype and validate its feasibility in a constructed rugged environment. For videos, simulation benchmarks, and code, please visit {\small \url{https://wyd0817.github.io/project-mrs-cwc/}}.

\end{abstract}

\section{INTRODUCTION}

\begin{figure}[t]
\begin{center}
\begin{minipage}[b]{.49\columnwidth}
\centering
\includegraphics[width=\columnwidth,height=3cm]{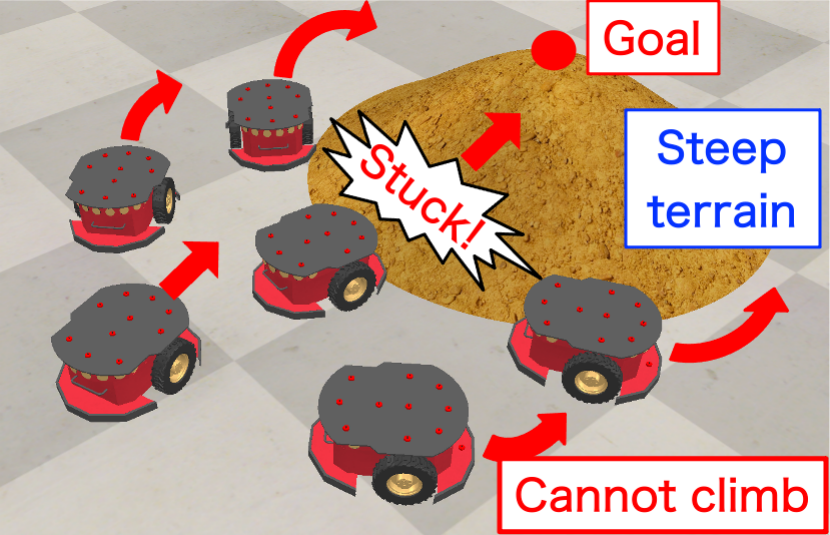}
\subcaption{Discrete MRS}
\end{minipage}
\begin{minipage}[b]{.49\columnwidth}
\centering
\includegraphics[width=\columnwidth,height=3cm]{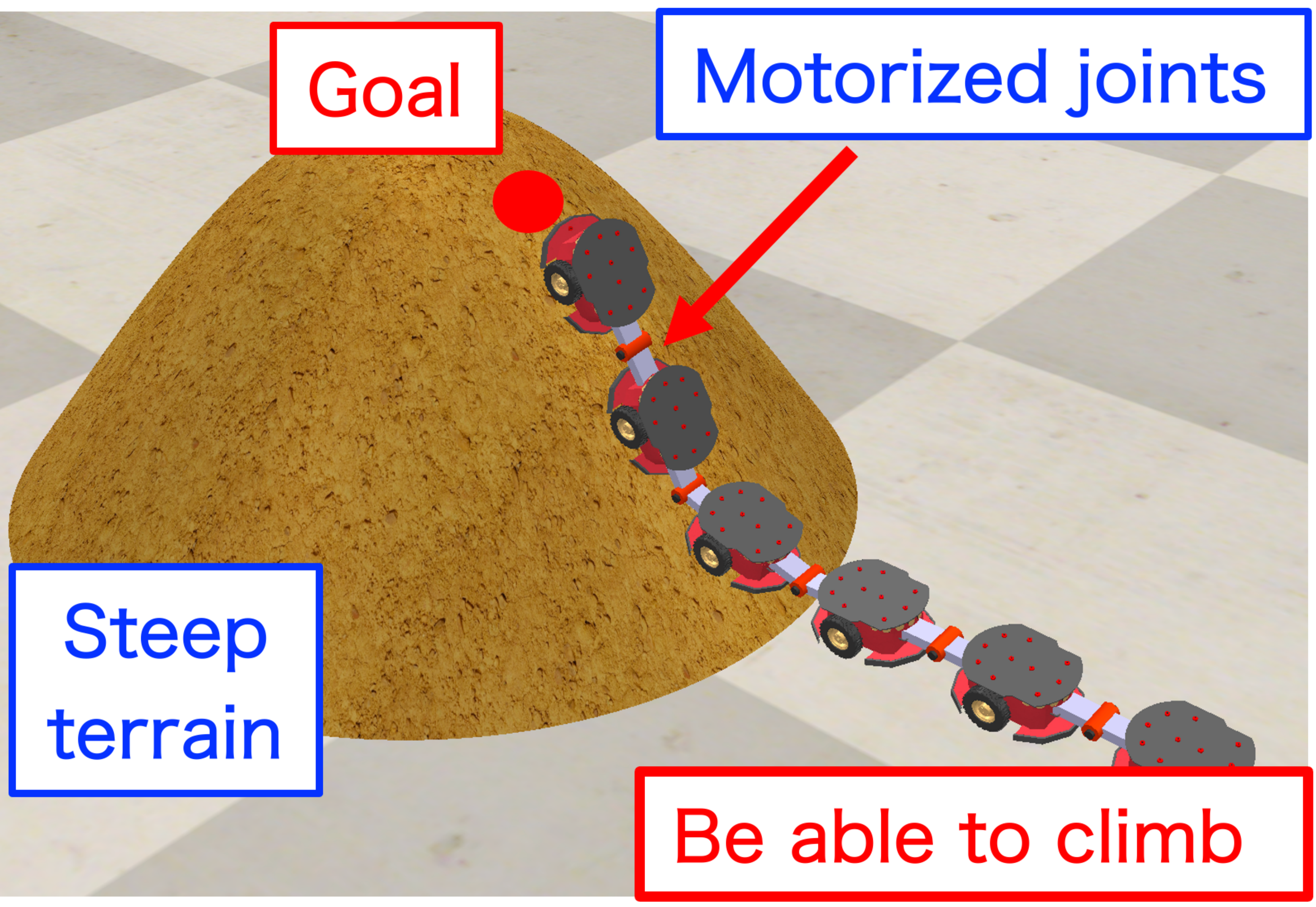}
\subcaption{Modular MRS}
\end{minipage}
\caption{\label{fig:bg}Traditional MRS in rugged environment navigation}
\end{center}
\end{figure}

As multi-robot systems (MRS) demonstrate advantages across applications, their navigation in 3D rugged terrains has become a key research focus in robotics. Traditional MRS navigation \cite{c1,c2,c3,c4} relies on mutual observation and communication, integrating environmental perception and path planning to coordinate movement and achieve group navigation in various environments. These discrete MRS offer high flexibility and adaptability. However, in rugged terrains such as mountainous or highly obstructed areas, individual mobility limitations often cause robots to become stuck, as shown in Fig.~\ref{fig:bg}(a), hindering effective navigation.

To enhance the mobility of MRS in rugged terrains, researchers have explored modular robotic systems \cite{c5,c6,c7,c8,c9}, which interconnect discrete robotic units through controllable joints with stiff links as shown in Fig.~\ref{fig:bg}(b). These systems have been proven effective in improving the traversability of MRS in challenging terrains compared to individual robotic units \cite{c5,c6,c7}. However, such “strongly constrained” systems still face challenges in navigating unknown rugged environments. For navigation, precise adaptation of a high-degree-of-freedom body to complex terrains requires sophisticated sensing and control systems for precise perception and path planning. For body control, coordinating highly redundant multi-joint structures increases system complexity and severely limits flexibility due to the strong coupling between motor motion control and compliance control, along with the inherent limitations of motor-based compliance control.

\begin{figure}[t]
\begin{center}
\begin{minipage}[b]{\columnwidth}
\centering
\includegraphics[width=\columnwidth]{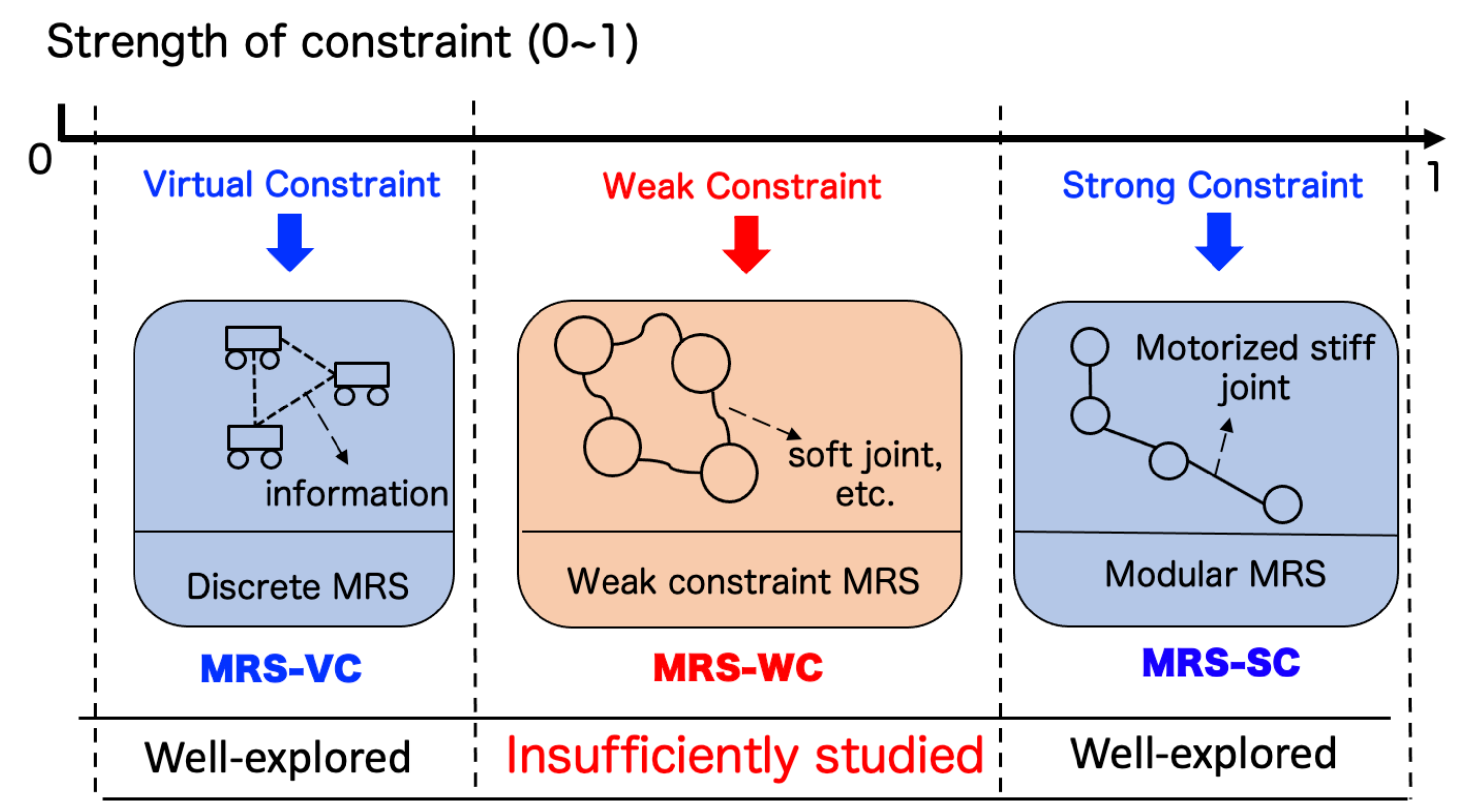}
\end{minipage}
\caption{\label{fig:self_pos}MRS classification by constraint strength, in ascending order: Virtual Constraint MRS (MRS-VC)—robotic units operate without physical constraints, relying solely on communication and coordinated motion; Weak Constraint MRS (MRS-WC)—physical constraints limit certain degrees of freedom while preserving some passive ones; Strong Constraint MRS (MRS-SC)—physical constraints completely restrict or precisely control all degrees of freedom.}
\end{center}
\end{figure}


To overcome the limitations of discrete and modular MRS in rugged terrain traversal and flexibility—without relying on complex computation and perception—this study introduces an intermediate system: the Weak Constraint Multi-Robot System (MRS-WC). MRS-WC leverages constraints such as flexible materials, passive joints, or contact friction to restrict certain degrees of freedom between robot units while preserving others. As shown in Fig.~\ref{fig:self_pos}, existing MRS and MRS-WC can be positioned along a 0–1 axis based on inter-robot constraint strength. Near 0, discrete MRS rely solely on virtual constraints from communication and coordination, without direct physical constraints; we refer to these as Virtual Constraint MRS (MRS-VC). Near 1, modular MRS feature high constraint levels through controllable joints or fastening elements, restricting or actively controlling all degrees of freedom; we term these Strong Constraint MRS (MRS-SC). The proposed MRS-WC lies between MRS-VC and MRS-SC, indicating that inter-robot constraints are partially restricted while some degrees of freedom remain free. This “selective partial constraint” strategy balances mobility and flexibility for rough terrain navigation while passively adapting in specific degrees of freedom, significantly reducing active control and sensing complexity.

Current MRS research primarily focuses on MRS-VC and MRS-SC, with relatively little exploration of MRS-WC. Representative MRS-WC implementations include the Kilobot Soft Robot by Pratissoli et al. \cite{c10}, which employs spring-connected kilobots to achieve coordinated motion with high individual error tolerance, and the boundary-constrained swarm robot by Karimi et al. \cite{c11,c12}, which utilizes flexible membranes filled with plastic particles. This system enables adaptive object grasping, system deformation, and target tracking in planar environments through granular jamming-induced rigidity modulation. Additionally, certain snake- and centipede-inspired robots can be classified as MRS-WC when considering their interconnected body segments as individual robotic units. Notable examples include the Genbu snake robot by Kimura et al. \cite{c13}, featuring elastic passive joints between two-wheeled modules; the Gunryu1 MRS by Hirose et al. \cite{c14}, which links independent tracked units via passive elastic arms; and the centipede-inspired i-CentiPot by Osuka et al. \cite{c15}, employing soft-axis actuation with elastic sponge inter-segment connections. These robots demonstrate a degree of rough terrain traversal capability.

However, existing MRS-WC systems still face critical limitations in 3D rough terrain navigation. The Kilobot Soft Robot \cite{c10} and boundary-constrained swarm robot \cite{c11,c12} are designed primarily for planar mobility and lack adaptations for effective movement in rugged terrains. While Genbu \cite{c13}, Gunryu1 \cite{c14}, and i-CentiPot \cite{c15} can traverse moderately rugged environments, they rely solely on passive compliance without trunk force, rendering them incapable of overcoming highly rugged terrains. Moreover, these studies primarily use passive materials to connect body segments, focusing on the design and control of individual robotic units rather than exploring weak constraint design and control. This highlights a key insight: a static degree-of-freedom design in weak constraints is insufficient for navigating complex and highly rugged environments. Instead, dynamically adjusting constraint stiffness in different directions over time—based on system state and environmental conditions—is crucial for enhancing adaptability to unknown terrains.


\begin{figure}[t]
    \centering
    \begin{minipage}[b]{\columnwidth}
        \centering
        \includegraphics[width=\columnwidth]{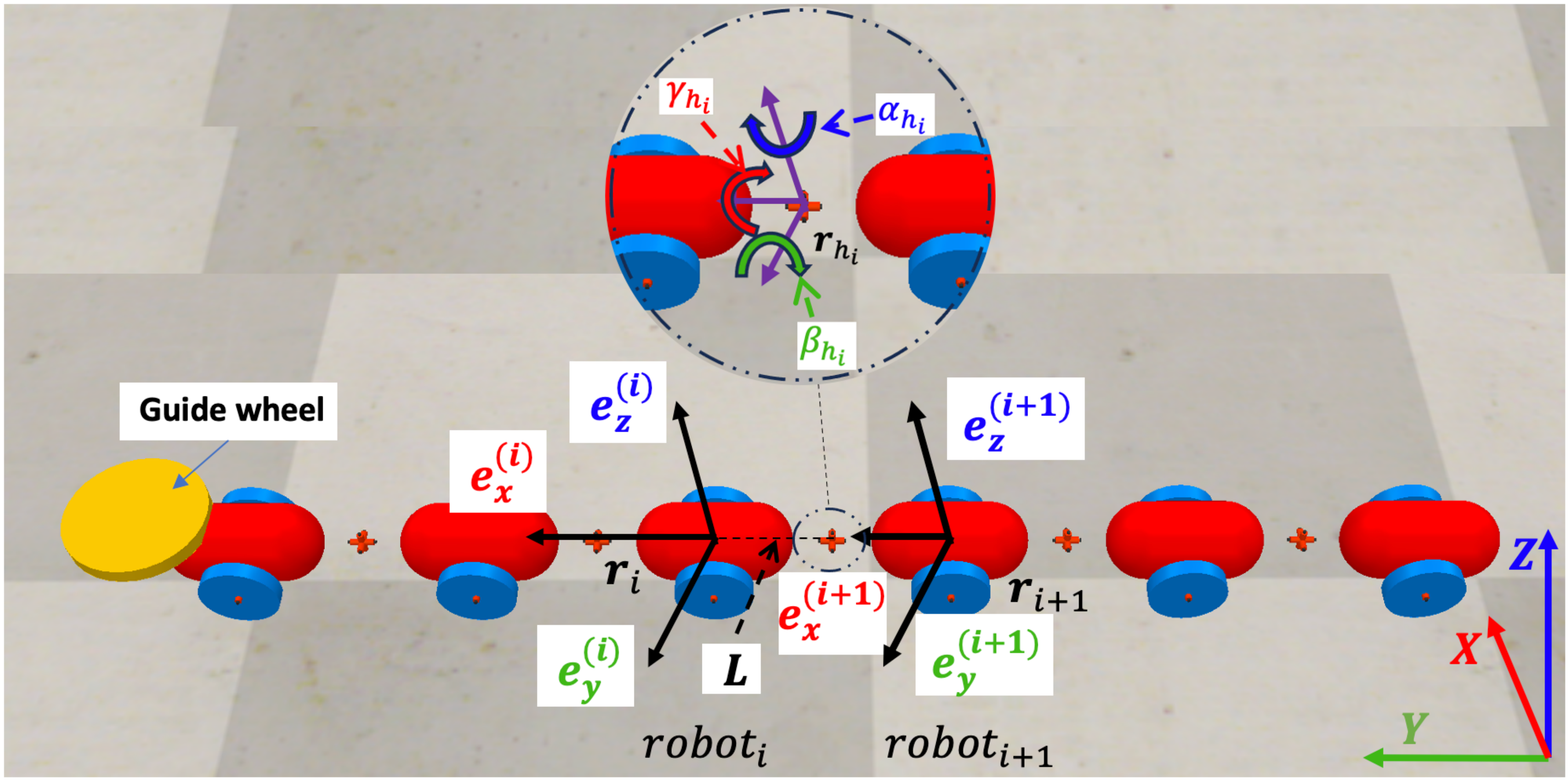}
        \subcaption{3D model}
    \end{minipage}
    \begin{minipage}[b]{\columnwidth}
        \centering
        \includegraphics[width=\columnwidth]{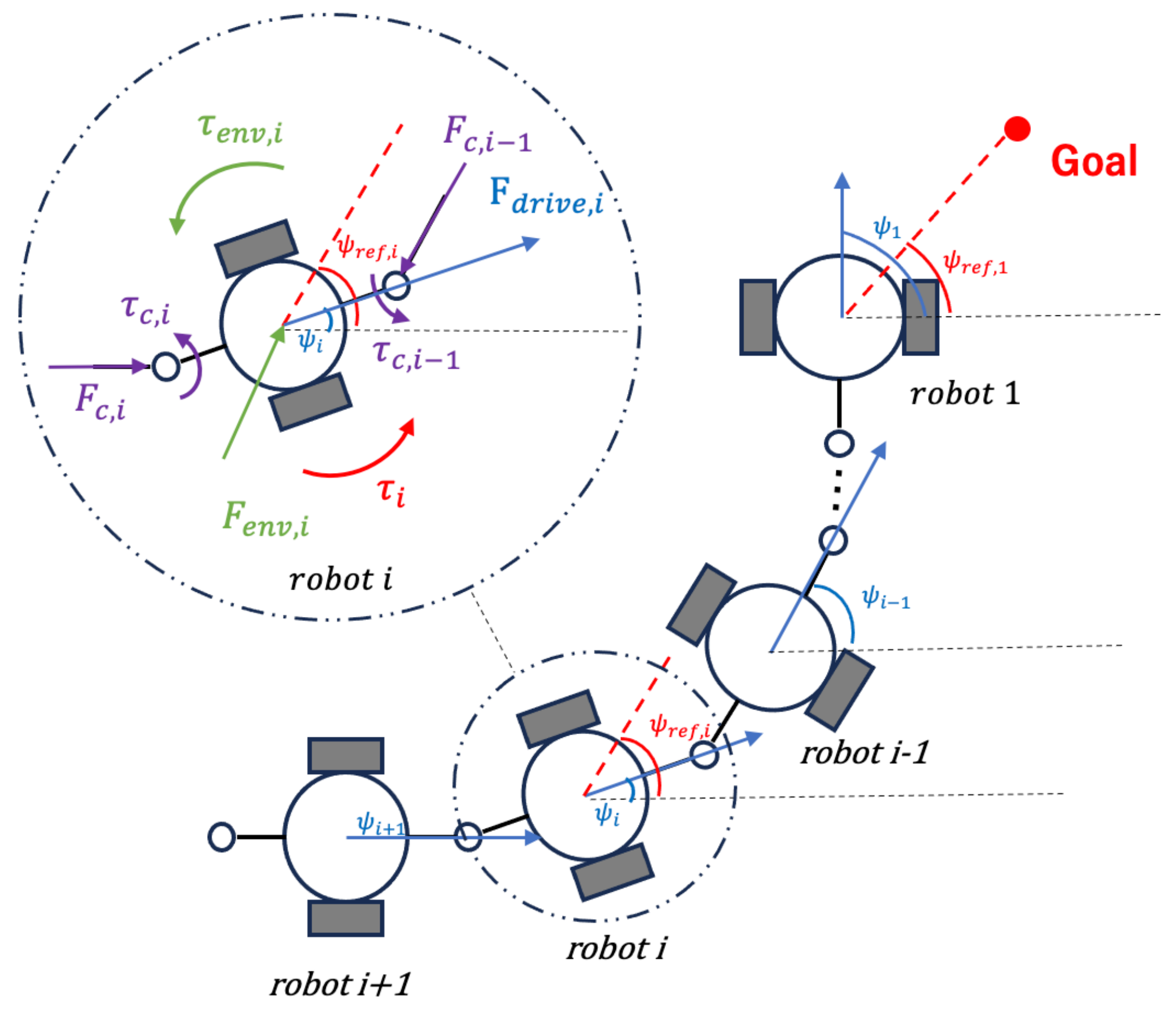}
        \subcaption{Simplified 2D model}
    \end{minipage}
    \caption{\label{fig:model_3}Theoretical modeling of the MRS-CWC}
\end{figure}
Building upon the aforementioned considerations, we propose a Multi-Robot System with Controllable Weak Constraint (MRS-CWC). The core features of MRS-CWC include:

\begin{itemize}
    \item \textbf{Dynamic Adaptation in Direct Interaction with the Environment:} By continuously modulating weak constraint stiffness in real-time interactions with the environment, the system balances flexibility and mobility, enabling dynamic compliance or resistance to external forces for effective terrain traversal.
\item \textbf{Minimal Perception and Computation Requirements:} Without relying on complex environmental perception or path planning, the system navigates rugged terrains by tracking the target direction and dynamically adjusting compliance or resistance to terrain variations.
\end{itemize}

This study makes the following key contributions:

\begin{enumerate}
\item Introduce the concepts of MRS-WC and MRS-CWC, along with a novel MRS classification based on constraint strength.
\item Develop an MRS-CWC system capable of navigating highly rugged terrains solely through target direction tracking and dynamic constraint stiffness adjustment.
\item Present a standardized simulation benchmark dataset with 100 diverse rugged terrain environments, categorized by average slope, providing a unified and reproducible platform for evaluating robotic navigation algorithms in rough terrains.
\end{enumerate}

The remainder of this paper is organized as follows: Section 2 presents the theoretical modeling of the MRS-CWC system, Section 3 describes the controller design, Section 4 covers simulation and experimental validation, and Section 5 concludes the study by summarizing the system’s advantages, limitations, and future research directions.

\section{Theoretical Modeling of the MRS-CWC System}


The MRS-CWC system can comprise multiple mobile robots in chain-like, networked, or more complex configurations. Here, we focus on a simple chain of six two-wheeled differential-drive robots (indexed \(i = 1,\dots,6\)) connected by three-DOF elastic joints, as shown in Fig.~\ref{fig:model_3} (a).

Each robot \(i\) carries a right-handed, orthonormal coordinate frame \(\{\mathbf{e}_x^{(i)}, \mathbf{e}_y^{(i)}, \mathbf{e}_z^{(i)}\}\) fixed to its body. The frame’s origin is \(\bm{r}_i = [x_i,\,y_i,\,z_i]^T\), expressed in the world coordinate system. The robot’s orientation relative to the world frame is given by roll, pitch, and yaw angles \(\phi_i,\theta_i,\psi_i\), which are rotations about \(\mathbf{e}_x^{(i)}, \mathbf{e}_y^{(i)}, \mathbf{e}_z^{(i)}\), respectively. The corresponding rotation matrix is \(\bm{R}_i = \bm{R}_z(\psi_i)\,\bm{R}_y(\theta_i)\,\bm{R}_x(\phi_i)\), where \(\bm{R}_x(\cdot)\), \(\bm{R}_y(\cdot)\), and \(\bm{R}_z(\cdot)\) are standard rotation matrices around the \(x\)-, \(y\)-, and \(z\)-axes.

\subsection{Definition and Mathematical Description of Constraints}

Similarly, the stiffness-controllable joint \(h_i\) connecting robots \(i\) and \(i+1\) has position \(\bm{r}_{h_i} = [x_{h_i},\,y_{h_i},\,z_{h_i}]^T\).

Firstly, the joint \(h_i\) enforces a constant distance \(L\) from each robot’s origin, leading to
\begin{equation}
\left\{
\begin{array}{l}
    \|\bm{r}_{h_i} - \bm{r}_i\|_2^2 = L^2,\\
    \|\bm{r}_{h_i} - \bm{r}_{i+1}\|_2^2 = L^2.
\end{array}
\right.
\end{equation}

Then, the joint also imposes rotational constraints described by  
\begin{equation}
\bm{\tau}_{h_i} = -\bm{K}_{\text{rot}, h_i}\,\bm{\Theta}_{h_i},
\label{eq:constraint_eq_2}
\end{equation}  
where \(\bm{\Theta}_{h_i} = [\alpha_{h_i},\,\beta_{h_i},\,\gamma_{h_i}]^T\) represents the rotation angles of joint \(h_i\) about \(\bm{e}_z^{(i)}, \bm{e}_y^{(i)},\) and \(\bm{e}_x^{(i)}\), respectively, with the zero position defined as the state shown in Fig.~\ref{fig:model_3}(a). The corresponding elastic restoring torques are given by \(\bm{\tau}_{h_i} = [\tau_{\alpha,h_i},\,\tau_{\beta,h_i},\,\tau_{\gamma,h_i}]^T\). The diagonal stiffness matrix is
\begin{equation}
\bm{K}_{\text{rot}, h_i} = \operatorname{diag}\bigl(K_{\alpha, h_i}(t),\, K_{\beta, h_i},\, K_{\gamma, h_i}\bigr),
\end{equation}
where \(K_{\beta, h_i}\) and \(K_{\gamma, h_i}\) are constants for terrain adaptability. The term \(K_{\alpha, h_i}(t)\) is time-varying and provides controllable stiffness about the \(\bm{e}_z^{(i)}\) axis.

\subsection{Dynamic Model}

For each robot \(i\), the rigid-body dynamics are
\begin{equation}
\begin{cases}
m_i\,\ddot{\bm{r}}_i = \bm{F}_i + \bm{F}_{c,i-1} + \bm{F}_{c,i} + \bm{F}_{\text{env},i},\\
\bm{I}_i\,\ddot{\bm{\Theta}}_i = \bm{\tau}_i + \bm{\tau}_{c,i-1} + \bm{\tau}_{c,i} + \bm{\tau}_{\text{env},i},
\end{cases}
\end{equation}
where \(m_i\) is the mass of robot \(i\), \(\ddot{\bm{r}}_i = [\ddot{x}_i,\,\ddot{y}_i,\,\ddot{z}_i]^T\) is its linear acceleration, and \(\bm{F}_i\) is the active driving force. The terms \(\bm{F}_{c,i-1}\) and \(\bm{F}_{c,i}\) are constraint forces from joints \(h_{i-1}\) and \(h_i\), while \(\bm{F}_{\text{env},i}\) accounts for environmental effects. The inertia matrix is \(\bm{I}_i = \operatorname{diag}(I_{x,i},\,I_{y,i},\,I_{z,i})\), and \(\ddot{\bm{\Theta}}_i = [\ddot{\phi}_i,\,\ddot{\theta}_i,\,\ddot{\psi}_i]^T\) is the angular acceleration. The active driving torque is \(\bm{\tau}_i\), while \(\bm{\tau}_{c,i-1}\) and \(\bm{\tau}_{c,i}\) come from adjacent joints, given by \(\bm{\tau}_{c,i} = \bm{K}_{\text{rot}}\,\bm{\Theta}_{h_i}\) following \eqref{eq:constraint_eq_2}. For boundary robots, \(\bm{F}_{c,i-1} = \bm{0}\) and \(\bm{\tau}_{c,i-1} = \bm{0}\) at \(i=1\), and \(\bm{F}_{c,i} = \bm{0}\) and \(\bm{\tau}_{c,i} = \bm{0}\) at \(i=6\).

Because motion control mainly occurs in the \(XY\) plane and stiffness control is primarily focused on the yaw direction—where other degrees of freedom adopt a passive strategy with fixed stiffness—the three-dimensional system is approximated as planar. As shown in Fig.~\ref{fig:model_3} (b), each robot’s pose is \(\bm{q}_i = [x_i,\,y_i,\,\psi_i]^T\). Each joint \(h_i\) is modeled as a hinge with one rotational DOF \(\alpha_{h_i} = \psi_i - \psi_{i+1}\). Thus, the simplified dynamics are
\begin{equation}\label{eq:dy_2d}
\mathbf{M}_i \,\ddot{\mathbf{q}}_i =
\mathbf{F}_{i} + \mathbf{F}_{c,i-1} + \mathbf{F}_{c,i} + \mathbf{F}_{\text{env},i},
\end{equation}
where \(\mathbf{M}_i = \operatorname{diag}(m_i, m_i, I_{z,i})\). The generalized force components include 
\begin{equation*}
\mathbf{F}_{i} =
\begin{bmatrix}
F_{\text{drive},i} \cos\psi_i \\
F_{\text{drive},i} \sin\psi_i \\
\tau_i
\end{bmatrix},
\quad
\mathbf{F}_{c,i} =
\begin{bmatrix}
F_{c,x,i} \\
F_{c,y,i} \\
K_{\alpha, h_{i}}(t)(\psi_{i+1} - \psi_i)
\end{bmatrix},
\end{equation*}

\begin{equation*}
\mathbf{F}_{c,i-1} =
\begin{bmatrix}
F_{c,x,i-1} \\
F_{c,y,i-1} \\
K_{\alpha, h_{i-1}}(t)(\psi_{i-1} - \psi_i)
\end{bmatrix},
\quad
\mathbf{F}_{\text{env},i} =
\begin{bmatrix}
F_{\text{env},x,i} \\
F_{\text{env},y,i} \\
\tau_{\text{env},i}
\end{bmatrix}.
\end{equation*}

Here, \(F_{\text{drive},i}\) is the forward driving force of robot \(i\), and \(\tau_i\) is its steering torque.
For boundary robots, \(\bm{F}_{c,i-1} = \bm{0}\) at \(i=1\), and \(\bm{F}_{c,i} = \bm{0}\) at \(i=6\).

\section{Control Model of the MRS-CWC System}
The MRS-CWC control model consists of two parts: motion control of the robot units and stiffness control for the weak constraints. The former governs vector~\(\mathbf{F}_{i}\) in \eqref{eq:dy_2d}, comprising the forward driving force \(\bm{F}_{\text{drive},i}\) and steering torque \(\tau_i\). The latter manages the controllable stiffness coefficient \(K_{\alpha,h_i}(t)\) in \(\mathbf{F}_{c,i}\).

\subsection{Motion Control Model for Robot Units}
A leader-follower scheme is employed. Robot~1 serves as the leader to track the target direction, determined by the target coordinates \(\bigl(x_{\text{target}}, y_{\text{target}}\bigr)\), while robots \(i=2,\dots,6\) follow the heading of their immediate predecessor. All robots maintain a constant forward speed \(v_0\). The controller is defined by
\begin{equation}
\left\{
\begin{array}{l}
v_{\text{forward},i} = v_0,\\
\dot{\psi}_i = K_p \bigl(\psi_{\text{ref},i} - \psi_i\bigr).
\end{array}
\right.
\end{equation}
where $v_{\text{forward},i}$ is the forward speed of robot~$i$, \(K_p\) is the heading tracking gain, \(\psi_i\) is the heading of robot~\(i\), and
\begin{equation}
\psi_{\text{ref},i} =
\begin{cases}
\arctan2\bigl(y_{\text{target}} - y_1,\; x_{\text{target}} - x_1\bigr), & i=1,\\
\psi_{i-1}, & i=2,\dots,6.
\end{cases}
\end{equation}

Each wheel motor uses a proportional (P) controller:
\begin{equation}
\left\{
\begin{aligned}
\tau_{L,i} &= K_m\bigl(\omega_{L,i} - \omega_{L_{\text{actual}},i}\bigr),\\
\tau_{R,i} &= K_m\bigl(\omega_{R,i} - \omega_{R_{\text{actual}},i}\bigr),
\end{aligned}
\right.
\end{equation}
where \(\tau_{L,i}\) and \(\tau_{R,i}\) are the left and right wheel torques, \(\omega_{L,i}\) and \(\omega_{R,i}\) are the commanded wheel speeds, \(K_m\) is the torque control gain, and \(\omega_{L_{\text{actual}},i}\), \(\omega_{R_{\text{actual}},i}\) are the measured wheel speeds.

Substituting these motor controls into the two-wheeled robot kinematics yields the forward driving force \(\mathbf{F}_{\text{drive},i}\) and steering torque \(\tau_i\):
\begin{equation}
\left\{
\begin{array}{l}
F_{\text{drive},i} = \frac{2K_m v_0}{r^2}
- \frac{K_m}{r}\Bigl(\omega_{L_{\text{actual}},i} + \omega_{R_{\text{actual}},i}\Bigr),\\
\tau_i = \frac{K_m d^2 K_p\bigl(\psi_{\text{ref},i} - \psi_i\bigr)}{2r^2}
- \frac{dK_m}{2r}\Bigl(\omega_{R_{\text{actual}},i} - \omega_{L_{\text{actual}},i}\Bigr).
\end{array}
\right.
\end{equation}
where \(r\) is the wheel radius and \(d\) is the distance between the wheels.

\subsection{Stiffness Control Model for Weak Constraints}
In the proposed MRS-CWC stiffness control system, all joints \( h_i \) share a uniformly regulated stiffness \( K_{\alpha}(t) \). Since the motion of the lead robot plays a dominant role in navigation, \( K_{\alpha}(t) \) is determined by the real-time state of robot 1 and its corresponding joint \( h_1 \). Subsequently, all other joints adopt this stiffness value. When \( K_{\alpha}(t) \) increases uniformly across all joints, the MRS aligns into a line configuration, maximizing forward propulsion for traversing steep terrain. Conversely, when \( K_{\alpha}(t) \) decreases uniformly, the system becomes more compliant, allowing it to yield to environmental forces and adapt to external guidance.

As one input to the \( K_{\alpha}(t) \) controller, we construct an estimator for the environmental torque \(\tau_{\text{env},i}\) acting on robot \(i\). From equation (5), the estimator is:
\begin{equation}
\hat{\tau}_{\text{env},i} = I_{z,i}\,\ddot{\psi}_i - \tau_i - K_{\psi}(t)\bigl(\psi_{i+1} - 2\psi_i + \psi_{i-1}\bigr).
\end{equation}
Since \( I_{z,i} \) is small, we approximate \( I_{z,i}\,\ddot{\psi}_i \approx 0 \). Knowing \(K_{\alpha}(t)\), and measuring \(\psi_{i+1}\), \(\psi_i\), \(\psi_{i-1}\), \(\tau_i\) via onboard sensors and inter-robot communication, we can compute \(\hat{\tau}_{\text{env},i}(t)\) in real time. Specifically, for robot 1:
\begin{equation}
\hat{\tau}_{\text{env},1} = -\tau_1 + K_{\alpha}(t)\bigl(\psi_1 - \psi_2\bigr).
\end{equation}

We then define:
\begin{equation}
K_{\alpha}(t) =
\begin{cases}
K_{\text{low}}, & \text{if } \tau_1 \cdot \tau_{\text{constraint},1} < 0 \text{ or } |\hat{\tau}_{\text{env},1}| \ge \tau_{\text{env,threshold}},\\[1mm]
K_{\text{high}}, & \text{otherwise},
\end{cases}
\end{equation}

where \(K_{\text{low}}\) and \(K_{\text{high}}\) are the two possible values of \(K_{\alpha}(t)\), corresponding to lower and higher stiffness, respectively. Under this scheme, if \(\tau_1\) and \(\tau_{\text{c},1}\) oppose each other—signaling that the constraint impedes active control if the environmental torque exceeds \(\tau_{\text{env,threshold}}\), we set \(K_{\alpha}(t)\) to \(K_{\text{low}}\) to mitigate adverse effects. Conversely, when \(\tau_1\) and \(\tau_{\text{constraint},1}\) align and the environmental torque remains below \(\tau_{\text{env,threshold}}\), \(K_{\psi}(t)\) is set to \(K_{\text{high}}\), enhancing steering effort in cooperation with active steering torque to traverse travel  through complex terrain.

It is important to note that, for both motion and stiffness control, the system only requires the target direction, each robot’s heading, and real-time wheel torques and speeds—no explicit terrain data are needed. Unlike conventional methods that rely on external sensing and path planning to avoid contact, this approach traverses rugged environments by simply following the target direction and adjusting joint stiffness. By directly interacting with the environment—sometimes yielding and sometimes resisting—it effectively achieves MRS navigation under challenging conditions.

\section{Simulation And Experiments}

\begin{figure}[t]
\begin{center}
\begin{minipage}[b]{.49\columnwidth}
\centering
\includegraphics[width=\columnwidth,height=3cm]{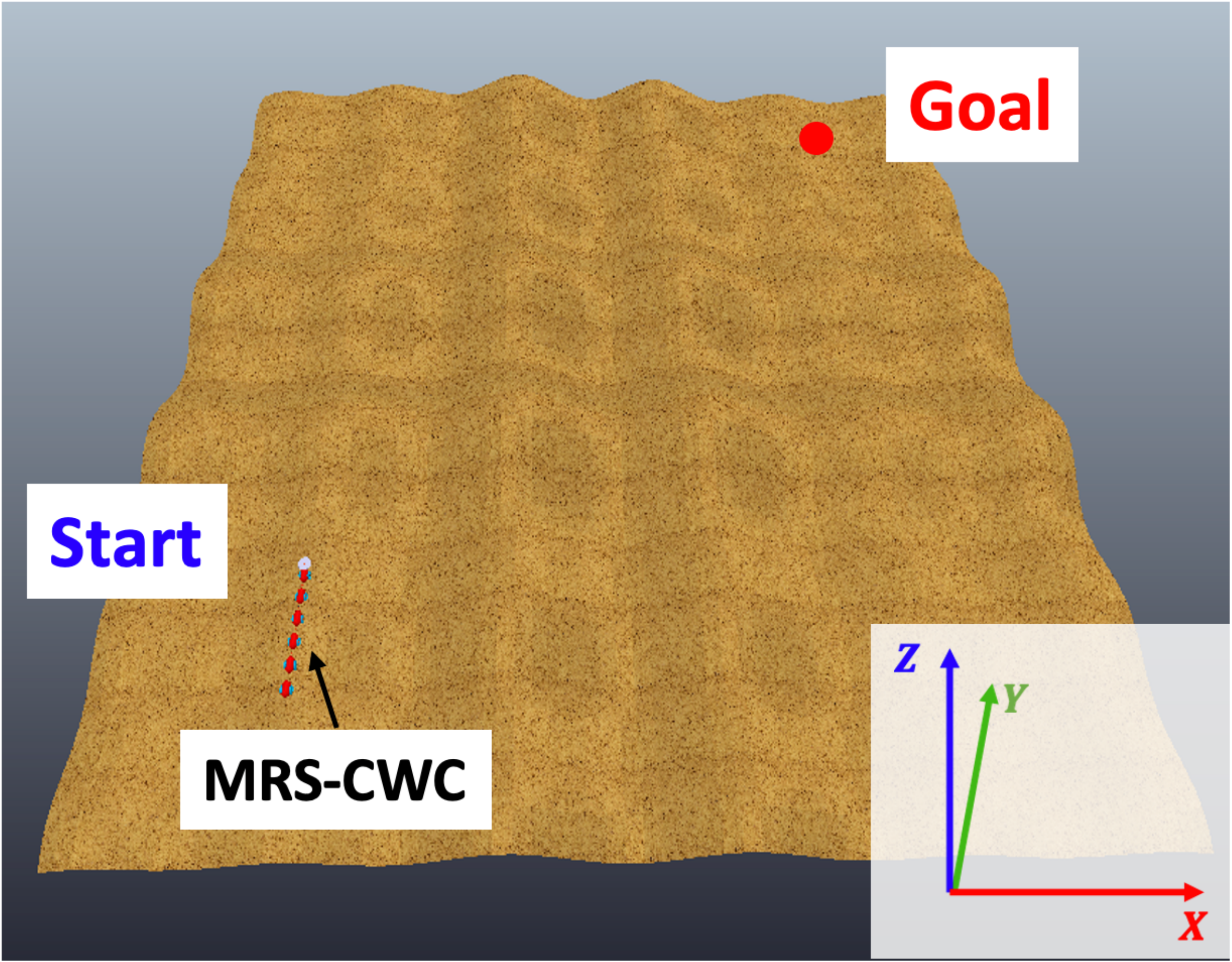}
\subcaption{One example in Group A}
\end{minipage}
\begin{minipage}[b]{.49\columnwidth}
\centering
\includegraphics[width=\columnwidth,height=3cm]{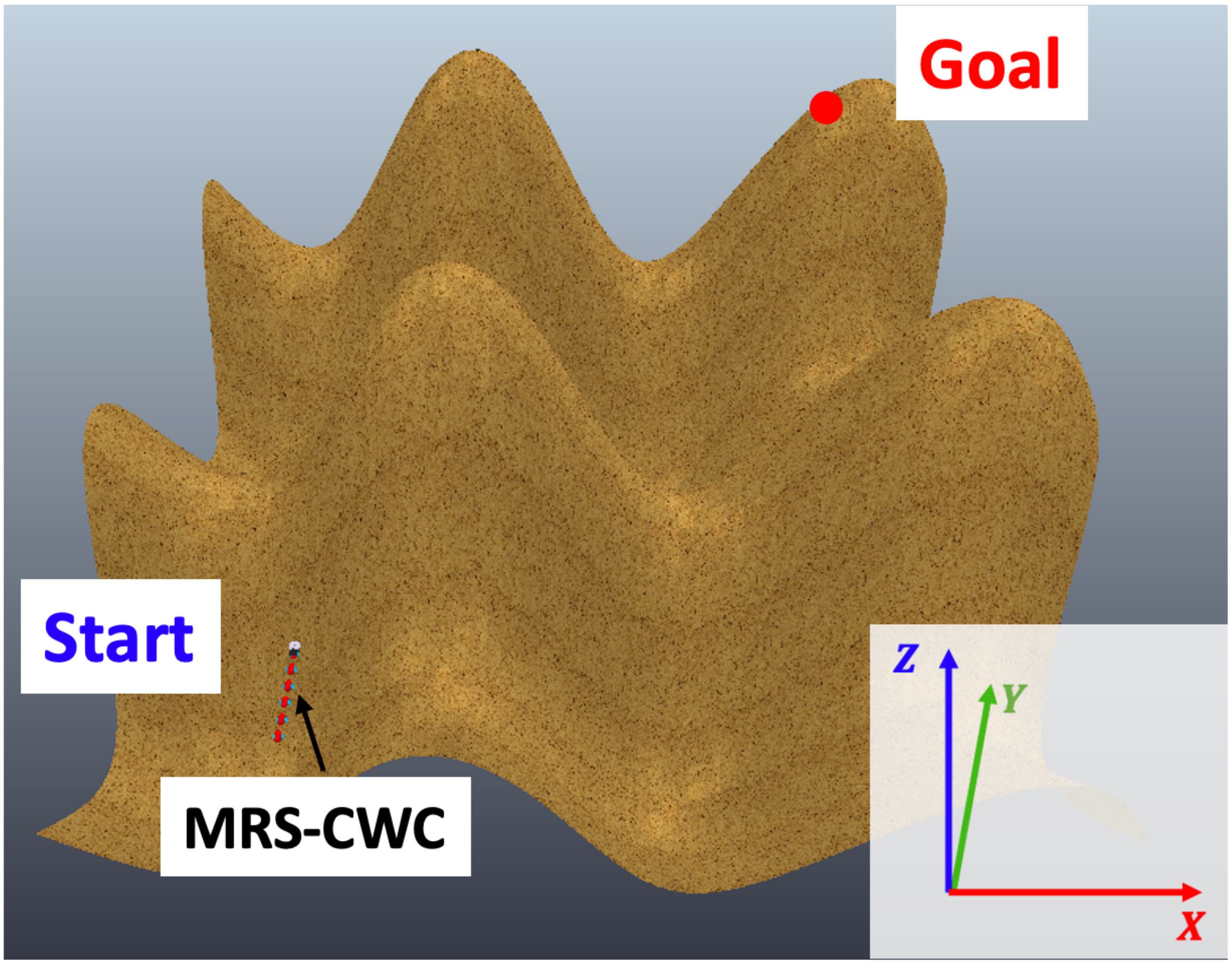}
\subcaption{One example in Group B}
\end{minipage}
\caption{\label{fig:sim_env}Simulation environment}
\end{center}
\end{figure}

\begin{table*}[htbp]
    \centering
    \renewcommand{\arraystretch}{1.3}
    \caption{Comparison Methods}
    \label{tab:MRS_variants}
    \begin{tabular}{|l|l|l|c|l|}
        \hline
        \textbf{Method} & \textbf{Constraint Type} & \textbf{Navigation Method} & \textbf{Env. Model} & \textbf{MRS Category} \\
        \hline
        \textbf{MRS-CWC + TDT (ours)}~†1 & Controllable Weak Constraint (CWC) & Target Direction Tracking (TDT) & None & MRS-WC \\
        \textbf{MRS-CWC + A\textsuperscript{*}} \cite{c19}~†2 & Controllable Weak Constraint (CWC) & A\textsuperscript{*} Path Planning & Global & MRS-WC \\
        \textbf{MRS-MJ + TDT} \cite{c16,c17}~†3 & Motorized Joint (MJ) & Target Direction Tracking (TDT) & None & MRS-SC \\
        \textbf{MRS-MJ + A\textsuperscript{*}} \cite{c19,c16,c17}~†4 & Motorized Joint (MJ) & A\textsuperscript{*} Path Planning & Global & MRS-SC \\
        \textbf{MRS-CS + TDT} \cite{c13}~†5 & Constantly Soft Joint (CS) & Target Direction Tracking (TDT) & None & MRS-WC \\
        \textbf{MRS-CS + A\textsuperscript{*}} \cite{c13,c19}~†6 & Constantly Soft Joint (CS) & A\textsuperscript{*} Path Planning & Global & MRS-WC \\
        \textbf{MRS-D + TDT} \cite{c18}~†7 & Discrete (D) & Target Direction Tracking (TDT) & None & MRS-VC \\
        \textbf{MRS-D + A\textsuperscript{*}} \cite{c19,c18}~†8 & Discrete (D) & A\textsuperscript{*} Path Planning & Global & MRS-VC \\
        \hline
    \end{tabular}

    \vspace{0.5em}

    \begin{minipage}{\textwidth}
        \footnotesize
        \raggedright
        \textbf{Notes:} \\ 
        \noindent †1 \textbf{MRS-CWC + TDT}: \textit{Proposed Method}, an MRS-CWC system using Target Direction Tracking (TDT). \\ 
        \noindent †2 \textbf{MRS-CWC + A\textsuperscript{*}}: \textit{Ablation Study}, replacing TDT of †1 with A\textsuperscript{*} path planning\cite{c19}, optimizing the trade-off between forward slope and tipping risk. \\ 
        \noindent †3 \textbf{MRS-MJ + TDT}: \textit{Baseline}, an MRS with motorized joints, following the ACM-R4 snake robot’s control method\cite{c16,c17} and TDT. \\ 
        \noindent †4 \textbf{MRS-MJ + A\textsuperscript{*}}: \textit{Baseline}, a variant of †3 using A\textsuperscript{*}\cite{c19} for path planning. \\ 
        \noindent †5 \textbf{MRS-CS + TDT}: \textit{Baseline}, an MRS with constantly soft joints, following the Genbu snake robot’s control method\cite{c13} and TDT navigation method. \\ 
        \noindent †6 \textbf{MRS-CS + A\textsuperscript{*}}: \textit{Baseline}, a variant of †5 using A\textsuperscript{*}\cite{c19} for path planning. \\ 
        \noindent †7 \textbf{MRS-D + TDT}: \textit{Baseline}, a discrete MRS using Follow-the-Leader control\cite{c18}. \\ 
        \noindent †8 \textbf{MRS-D + A\textsuperscript{*}}: \textit{Baseline}, a variant of †7 using A\textsuperscript{*}\cite{c19} for path planning. \\ 
    \end{minipage}

\end{table*}
To evaluate the effectiveness of the proposed method and its advantages over existing approaches, we conducted comparative simulation experiments using a CoppeliaSim-based benchmark dataset comprising 100 terrains of varying ruggedness. Additionally, we developed a physical prototype and validated its feasibility in a constructed rugged terrain.
\subsection{Simulation Experimental Setup}
The simulation environments were procedurally generated, drawing inspiration from the terrain design approach in CoppeliaSim’s official model library, using the equation:
\begin{equation}
\resizebox{\linewidth}{!}{$
H(x, y) = \left[ \sum_{i=1}^{3} w_i A_i \left( \sin(2\pi P_i + f_i x) + \sin(2\pi P_i + f_i y) \right) \right] dzs,
$}
\end{equation}
where the parameters \( A_i, P_i, f_i \) (\( i \in \{1,2,3\} \)) and $dzs$ were randomly assigned. The weighting coefficients were set to \( w_1 = 1 \), \( w_2 = 0.5 \), and \( w_3 = 0.1 \).
Using the average slope as a criterion, we selected two sets of 50 terrain samples:
\begin{itemize}
    \item \textbf{Group A} (average slope \(< 30^\circ\)): Moderate roughness.
    \item \textbf{Group B} (average slope \(> 30^\circ\)): Challenging terrain.
\end{itemize}

To avoid extreme scenarios, the maximum slope of both groups was constrained to 70° in all environments. Representative terrain samples from both groups are shown in Fig.~\ref{fig:sim_env}.

\subsection{Comparison Methods}

We evaluated eight methods, including the proposed approach, an ablation variant, and six baselines. For the proposed approach, parameters were set as \(K_{\text{low}} = 5\), \(K_{\text{high}} = 100\), \(v_0 = 0.075\,\text{m/s}\), \(K_p = 0.5\), \(\tau_{\text{env,threshold}} = 20\) Nm, \(K_{\beta,h_i} = 15\), and \(K_{\gamma,h_i} = 30\) for \(i \in \{1, \dots, 6\}\). Table~\ref{tab:MRS_variants} summarizes the characteristics of each method.

Each method was evaluated in both terrain groups, where the MRS had to navigate from the starting point \([0, 0, H(0, 0)]\) to the target \([9, 9, H(9, 9)]\). A trial was considered successful if the robot's XY-plane projection distance to the target fell below 0.5 m within 1000 seconds.

\subsection{Evaluation Metrics}
The following metrics were used to assess navigation performance in the simulation trials:
\begin{itemize}
    \item \textbf{Navigation Success Rate (NSR):} For 50 navigation trials in a given group, if the number of successful trials is \(N_s\), then
    \begin{equation}
    \text{NSR} = \frac{N_s}{50}.
    \end{equation}
    \item \textbf{Navigation Completion Rate (NCR):} Let \(L\) be the horizontal distance from the start to the target, and \(d\) be the minimum horizontal distance between the leader robot and the target during navigation. The effective navigation distance is defined as
    \begin{equation}
    D_{\text{ef}} = L - d,
    \end{equation}
    and the completion rate for each trial is
    \begin{equation}
    \text{NCR} = \frac{L - d}{L}.
    \end{equation}
    The average NCR over 50 trials in a group is denoted as \(\overline{\text{NCR}}\).
    \item \textbf{Navigation Efficiency (NEF):} If \(J\) is the actual journey the MRS traveled, then
    \begin{equation}
    \text{NEF} = \frac{D_{\text{ef}}}{J}.
    \end{equation}
    This metric represents the effective progress per unit journey traveled. The average NEF over 50 trials is denoted as \(\overline{\text{NEF}}\).
    \item \textbf{Navigation Energy Consumption (NEC):} By recording the wheel torques \(\tau_{L,i}\), \(\tau_{R,i}\) and speeds  \(\omega_{L,i}\), \(\omega_{R,i}\) of each unit in MRS over the navigation period, the total energy consumption \(E\) is computed as
    \begin{equation}
    E = \int_{0}^{T} \sum_{i=1}^{6} \bigl(\left |\tau_{L,i}\,\omega_{L,i} \right | +\left |\tau_{R,i}\,\omega_{R,i}\right | \bigr)\,dt.
    \end{equation}
    where T denotes the total navigation time. The NEC is then defined as
    \begin{equation}
    \text{NEC} = \frac{E}{D_{\text{ef}}},
    \end{equation}
    representing the energy consumption per unit effective navigation distance. The average NEC over 50 trials in a group is denoted as \(\overline{\text{NEC}}\).
\end{itemize}

\subsection{Simulation Results}

\begin{figure}[t]
    \centering
    \begin{minipage}[b]{\columnwidth}
        \centering
        \includegraphics[width=\columnwidth]{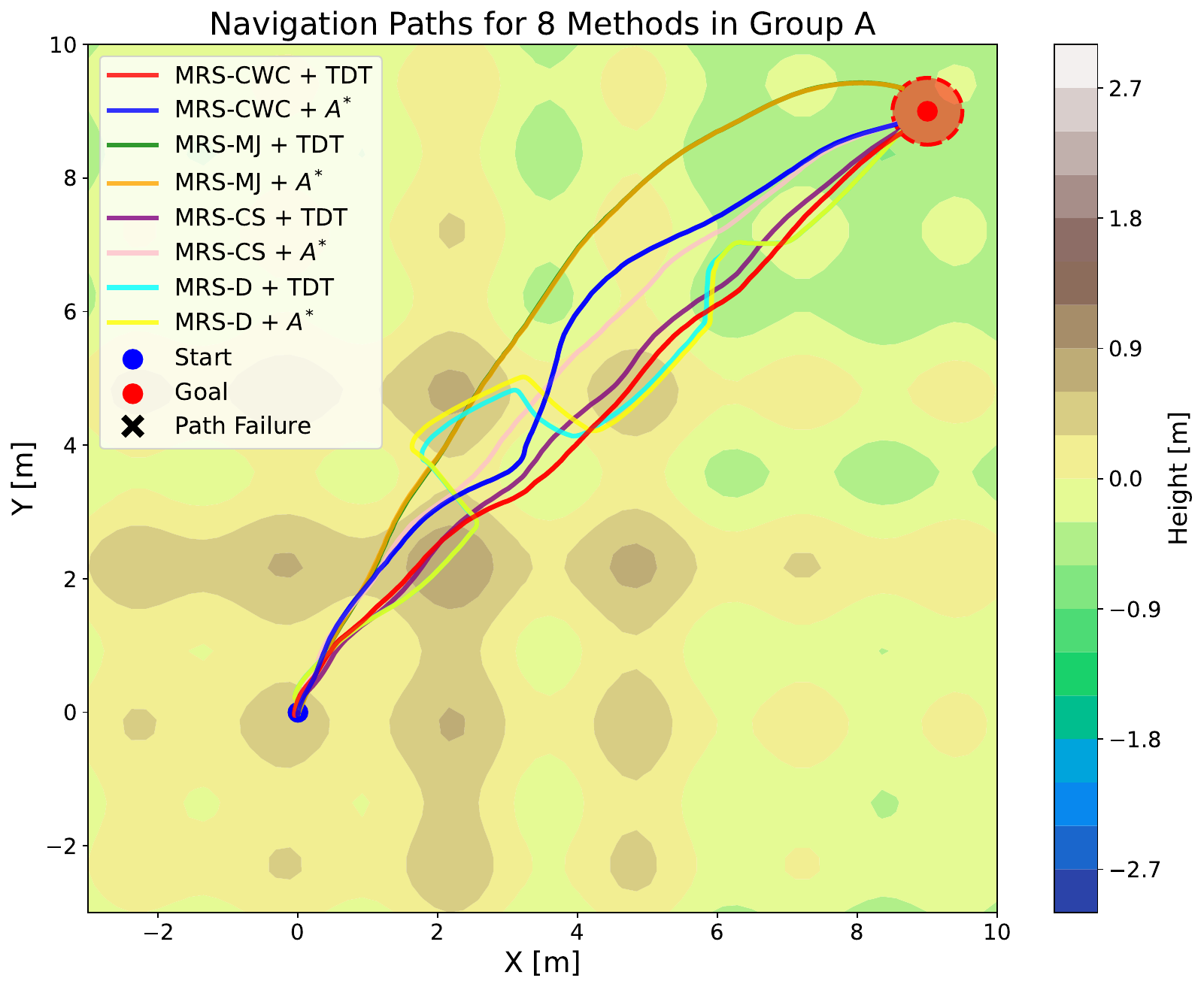}
    \end{minipage}
    \caption{\label{fig:rs_1}One example navigation results in Group A }
\end{figure}

\begin{figure}[t]
    \centering
    \begin{minipage}[b]{\columnwidth}
        \centering
        \includegraphics[width=\columnwidth]{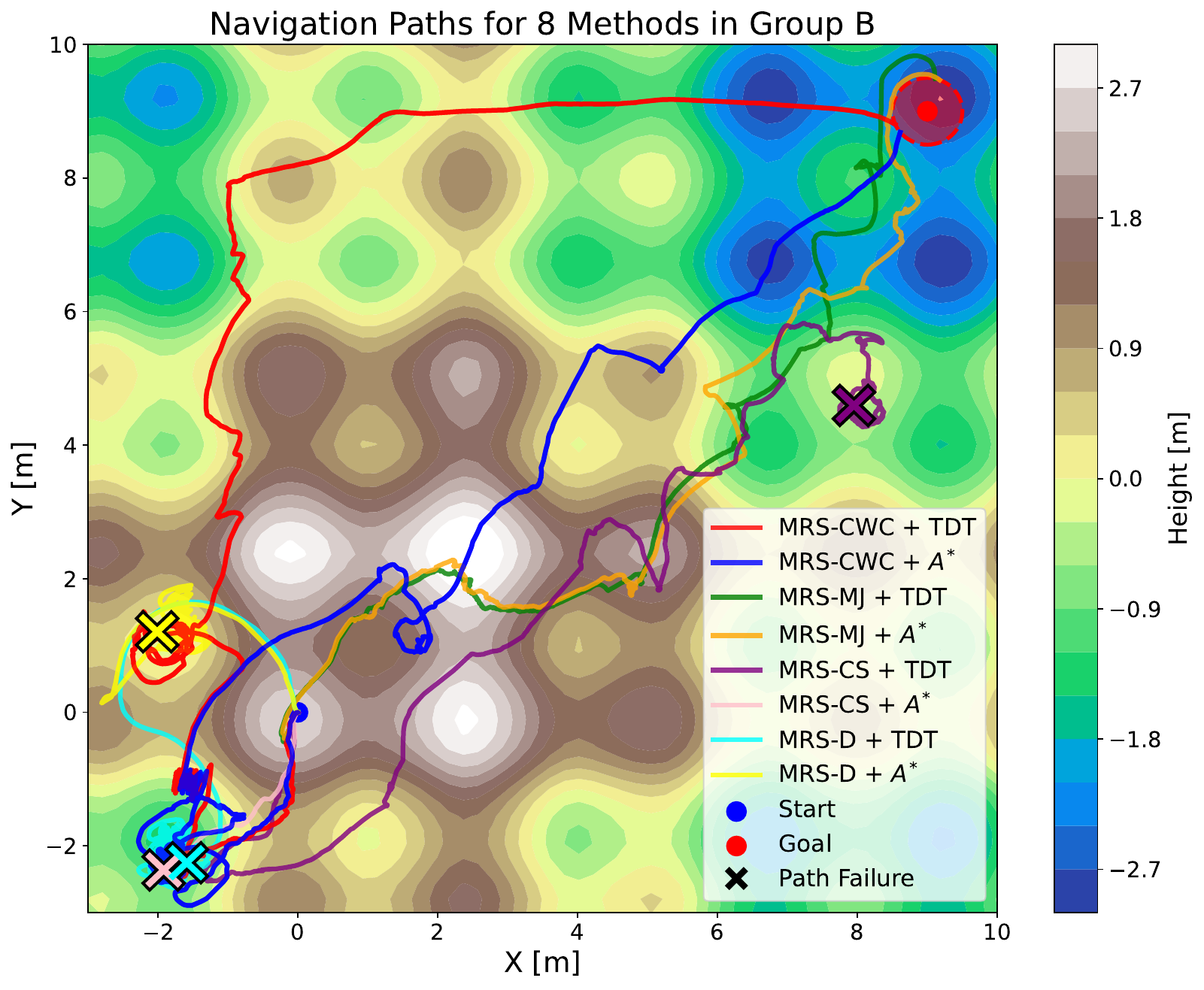}
    \end{minipage}
    \caption{\label{fig:rs_2}One example navigation results in Group B}
\end{figure}

In Fig.~\ref{fig:rs_1} and Fig.~\ref{fig:rs_2}, we present one example navigation results in Group A and Group B environments, while Table~\ref{tab:table1} and Table~\ref{tab:table2} summarize the evaluation metrics for each group. The data show that constraint mechanisms strongly influence multi-robot navigation in challenging terrains: in Group A, the six methods using MRS-CWC, MRS-MJ, and MRS-CS constraints perform comparably, with NSR, $\overline{\text{NCR}}$, $\overline{\text{NEF}}$, and $\overline{\text{NEC}}$ ranging from 92\%-96\%, 93\%-99\%, 0.70-0.78, and 258-278 J/m, respectively. In contrast, the two approaches based on MRS-D are notably weaker, reaching only 62\%-64\%, 79\%-81\%, 0.53-0.54, and 286-409 J/m. In Group B, where terrain ruggedness increases, MRS-CWC continues to lead (NSR 74\%-78\%, $\overline{\text{NCR}}$ 88\%-91\%, $\overline{\text{NEF}}$ 0.33-0.38, $\overline{\text{NEC}}$ 639-755 J/m). In contrast, MRS-MJ and MRS-CS demonstrate relatively lower performance (NSR 38\%-52\%, $\overline{\text{NCR}}$ 67\%-74\%, $\overline{\text{NEF}}$ 0.20-0.28, $\overline{\text{NEC}}$ 1797-3332 J/m), though they still outperform MRS-D, which fails completely (NSR 0\%, $\overline{\text{NCR}}$ 20\%-21\%, $\overline{\text{NEF}}$ 0.04-0.06, $\overline{\text{NEC}}$ 6000-9000 J/m).
 When comparing TDT and A\textsuperscript{*} under the same constraints, A\textsuperscript{*} slightly surpasses TDT, yet the difference is minimal, and MRS-CWC+TDT and MRS-CWC+A\textsuperscript{*} yield almost identical outcomes.

\begin{table}[t]
  \centering
  \caption{Performance Metrics for Methods in Group A}
  \label{tab:table1}
  \renewcommand{\arraystretch}{1.2} 
  \setlength{\tabcolsep}{3pt} 
  \begin{tabular}{lcccc}
    \toprule
    Method              & NSR (\%) ↑    & $\overline{\text{NCR}}$ (\%) ↑    & $\overline{\text{NEF}}$ ↑   & $\overline{\text{NEC}}$ (J/m) ↓ \\
    \midrule
    \textbf{MRS-CWC + TDT}       & \textbf{96.0} & \textbf{98.2} & \textbf{0.775}  & 269.0     \\
    MRS-CWC + $A^{*}$    & 94.0          & 97.5          & 0.755           & 277.1               \\
    MRS-MJ + TDT        & 94.0          & 97.7          & 0.701           & 267.6               \\
    MRS-MJ + $A^{*}$     & 92.0          & 97.4          & 0.707           & \textbf{258.9}     \\
    MRS-CS + TDT        & 88.0          & 93.4          & 0.723           & 261.8               \\
    MRS-CS + $A^{*}$     & 94.0          & 96.4          & 0.753           & 273.6               \\
    MRS-D + TDT         & 62.0          & 79.1          & 0.531           & 286.0               \\
    MRS-D + $A^{*}$      & 64.0          & 80.1          & 0.539           & 409.0               \\
    \bottomrule
  \end{tabular}
\end{table}

\begin{table}[t]
  \centering
  \caption{Performance Metrics for Methods in Group B}
  \label{tab:table2}
  \renewcommand{\arraystretch}{1.2} 
  \setlength{\tabcolsep}{3pt} 
  \begin{tabular}{lcccc}
    \toprule
    Method              & NSR (\%) ↑   & $\overline{\text{NCR}}$ (\%) ↑   & $\overline{\text{NEF}}$ ↑   & $\overline{\text{NEC}}$ (J/m) ↓ \\
    \midrule
    \textbf{MRS-CWC + TDT}       & 74.0  & \textbf{90.3}  & 0.336  & 754.5     \\
    MRS-CWC + $A^{*}$   & \textbf{78.0}  & 88.6  & \textbf{0.371}  & \textbf{639.7}     \\
    MRS-MJ + TDT        & 38.0  & 67.8  & 0.200  & 3332.5     \\
    MRS-MJ + $A^{*}$    & 52.0  & 73.2  & 0.248  & 2711.3     \\
    MRS-CS + TDT        & 40.0  & 73.8  & 0.253  & 1797.3     \\
    MRS-CS + $A^{*}$    & 48.0  & 69.5  & 0.273  & 2340.5     \\
    MRS-D + TDT         & 0.0   & 20.4  & 0.054  & 6530.0     \\
    MRS-D + $A^{*}$     & 0.0   & 20.6  & 0.046  & 8909.7     \\
    \bottomrule
  \end{tabular}
\end{table}

\subsection{Analysis}

Among the different MRS constraints, MRS-D performs well on flat terrain but struggles in rugged environments due to the absence of physical connections, making it prone to tipping or entrapment and significantly limiting its mobility. In contrast, MRS-MJ and MRS-CS benefit from physical constraints, enabling efficient traversal in moderately rugged terrains (Group A). However, in highly rugged terrains (Group B), MRS-MJ’s head-following body control algorithm generates wave-like motion when climbing steep slopes, dispersing propulsion and reducing climbing efficiency. Additionally, errors in joint torque sensors and delays in motor compliance control further limit its flexibility, negatively impacting overall navigation performance. MRS-CS, with its soft constraints, always passively conforms to terrain, leading to excessive deformation and side-slipping on steep slopes, often resulting in traversal failure. MRS-CWC overcomes these limitations by dynamically adjusting stiffness, balancing mobility and flexibility. When encountering extreme terrain or beneficial external forces, it softens to conform to obstacles, enabling smooth passage. Conversely, on traversable slopes, it stiffens to align units linearly, concentrating propulsion and enhancing mobility. This adaptive mechanism allows MRS-CWC to navigate rough terrains more effectively than all other configurations.

Regarding navigation strategies, MRS with A\textsuperscript{*} exhibited slightly better performance than TDT, but the difference was minimal. In particular, MRS-CWC+TDT and MRS-CWC+A\textsuperscript{*} achieved nearly identical results. This is because MRS with A\textsuperscript{*} frequently deviates from pre-planned paths in steep terrains, struggling to realign and reducing its effectiveness to a level similar to TDT. On the other hand, TDT operates without mapping, path planning, or additional sensors, significantly reducing system complexity and cost compared to A\textsuperscript{*} method.

Therefore, considering both performance and complexity, MRS-CWC+TDT emerges as the most well-rounded approach, achieving a favorable trade-off between mobility and flexibility while maintaining low system complexity and cost. It outperforms all eight compared methods.



\subsection{Prototype validation experiment}

\begin{figure}[t]
\begin{center}
\begin{minipage}[b]{\columnwidth}
\centering
\includegraphics[width=\columnwidth]{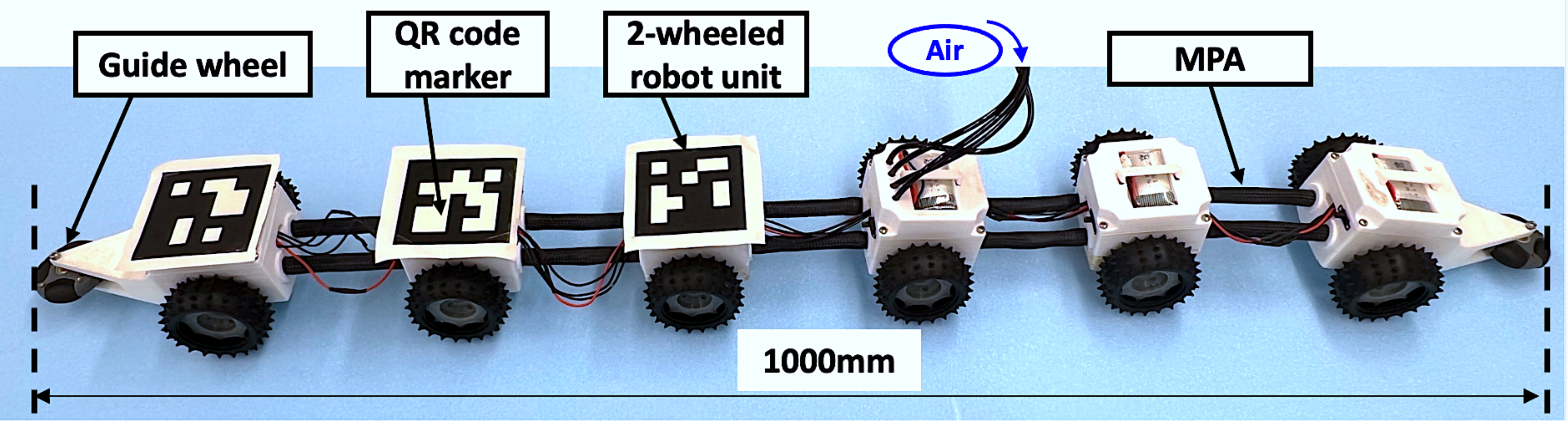}
\end{minipage}
\caption{\label{fig:prt}Prototype of MRS-CWC. We dynamically regulate the stiffness of constraints between robot units by controlling the inflation and deflation of two parallel connected MPAs.}
\end{center}
\end{figure}

As shown in Fig.~\ref{fig:prt}, we designed an MRS-CWC prototype with 6 independently controlled robotic units connected by two parallel rows of McKibben Pneumatic Artificial Muscles(MPA). All MPAs share a common air circuit, driven by an air pump. When inflated to 0.5 mPA, the dual-row MPAs increase yaw stiffness while maintaining pitch and roll flexibility to some extends for terrain adaptation. When deflated, all constraints become passively flexible.

\begin{figure}[t]
\begin{center}
\begin{minipage}[b]{\columnwidth}
\centering
\includegraphics[width=\columnwidth]{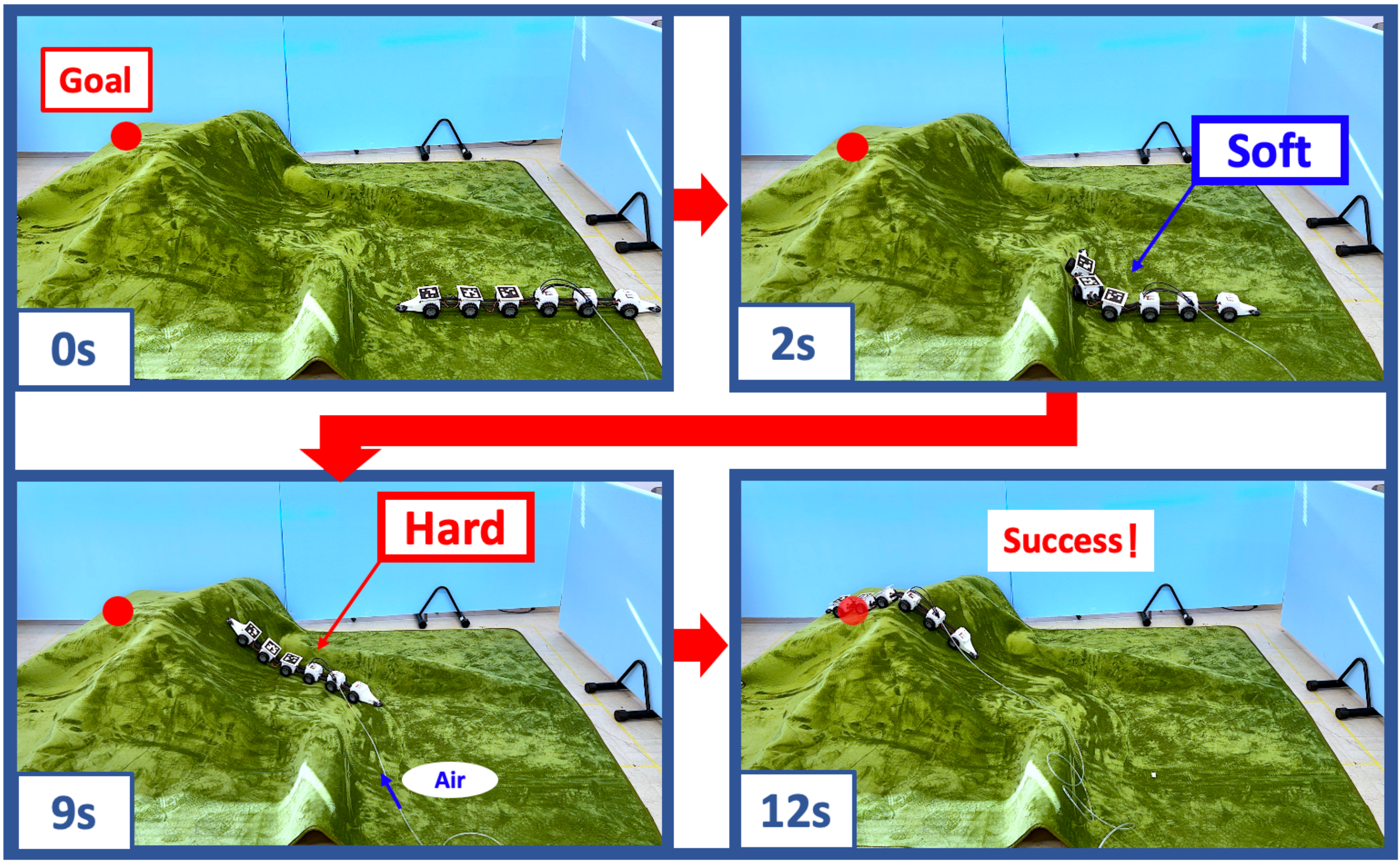}
\end{minipage}
\caption{\label{fig:exp}Validation experiment. The prototype stays soft under beneficial or irresistible external forces, stiffens to prevent slippage on traversable slopes, and navigates the terrain.}
\end{center}
\end{figure}

We conducted validation experiments in a rugged lab environment, as shown in Fig.~\ref{fig:exp}. The MRS-CWC prototype remains soft under beneficial or irresistible external forces, ensuring adaptive yielding. When climbing steep traversal slopes, MPAs stiffen the body to prevent slippage, enabling the system to cross the incline and reach the target. The results preliminarily validate the proposed method.

\section{Conclusion}
This study presents an MRS-CWC system that enables adaptive navigation in rugged terrains by dynamically adjusting constraint stiffness between two-wheeled robotic units, effectively balancing mobility and adaptability while maintaining low computational and observational costs. However, the current implementation depends on a uniformed stiffness control mechanism via the head module, which is effective for small-scale systems but necessitates a decentralized approach for larger robotic teams. Additionally, in scenarios containing untouchable obstacles, integrating lightweight perception modules for obstacle recognition is essential. Future work will focus on decentralized control strategies and perception integration to improve scalability and robustness in complex environments.


\addtolength{\textheight}{-12cm}   



\section*{ACKNOWLEDGMENT}

This research was supported in part by grants-in-aid for JSPS KAKENHI Grant Number JP22K14277, Grant-in-Aid for JSPS Fellows Grant Number 23KJ1445 and JST Moon-shot Research and Development Program JPMJMS2032 (Innovation in Construction of Infrastructure with Cooperative AI and Multi-Robots Adapting to Various Environments).



\begin{thebibliography}{99}

\bibitem{c1} E. Olcay, F. Schuhmann, and B. Lohmann, ``Collective navigation of a multi-robot system in an unknown environment,'' in Robotics and Autonomous Systems, vol. 132, Elsevier, 2020, p. 103--604. 
\bibitem{c2} A. T. Hayes and P. Dormiani-Tabatabaei, ``Self-organized flocking with agent failure: Off-line optimization and demonstration with real robots,'' in Proceedings of the 2002 IEEE International Conference on Robotics and Automation, vol. 4, IEEE, 2002, pp. 3900--3905.
\bibitem{c3} L. Silva Junior and N. Nedjah, ``Efficient strategy for collective navigation control in swarm robotics,'' in Procedia Computer Science, vol. 80, Elsevier, 2016, pp. 814--823.
\bibitem{c4}A.-R. Merheb, V. Gazi, and N. Sezer-Uzol, ``Implementation studies of robot swarm navigation using potential functions and panel methods,'' in IEEE/ASME Transactions on Mechatronics, vol. 21, no. 5, P. Editor, Ed. New York: IEEE/ASME, 2016, pp. 2556--2567.
\bibitem{c5} R. O’Grady, et al., ``Self-assembly on demand in a group of physical autonomous mobile robots navigating rough terrain,'' in Advances in Artificial Life, 8th European Conference, ECAL 2005, P. Editor, Ed. Berlin: Springer, 2005.
\bibitem{c6} Y. Ozkan-Aydin and D. I. Goldman, ``Self-reconfigurable multilegged robot swarms collectively accomplish challenging terradynamic tasks,'' in Science Robotics, vol. 6, P. Editor, Ed. New York: AAAS, 2021, p. eabf1628.
\bibitem{c7}S. Mizunuma, K. Motomura, and S. Hirose, ``Development of the arm-wheel hybrid robot ‘Souki-II’ (Total system design and basic components),'' in Intelligent Robots and Systems, 2009 IEEE/RSJ International Conference, P. Editor, Ed. New York: IEEE, 2009.
\bibitem{c8} L. Pfotzer, et al., ``KAIRO 3: A modular reconfigurable robot for search and rescue field missions,'' in Robotics and Biomimetics, 2014 IEEE International Conference, P. Editor, Ed. New York: IEEE, 2014.
\bibitem{c9}G. Granosik, M. G. Hansen, and J. Borenstein, ``The OmniTread serpentine robot for industrial inspection and surveillance,'' in Industrial Robot: An International Journal, vol. 32, no. 2, P. Editor, Ed. New York: Emerald, 2005, pp. 139--148.
\bibitem{c10}F. Pratissoli, et al., ``Coherent movement of error-prone individuals through mechanical coupling,'' in Nature Communications, vol. 14, no. 1, P. Editor, Ed. New York: Nature, 2023, p. 4063.
\bibitem{c11}M. A. Karimi, et al., ``A boundary-constrained swarm robot with granular jamming,'' in Soft Robotics, 2020 3rd IEEE International Conference, P. Editor, Ed. New York: IEEE, 2020.
\bibitem{c12}M. A. Karimi, et al., ``A self-reconfigurable variable-stiffness soft robot based on boundary-constrained modular units,'' in IEEE Transactions on Robotics, vol. 38, no. 2, P. Editor, Ed. New York: IEEE, 2021, pp. 810--821.
\bibitem{c13}H. Kimura and S. Hirose, ``Development of Genbu: Active wheel passive joint articulated mobile robot,'' in IEEE/RSJ International Conference on Intelligent Robots and Systems, vol. 1, IEEE, Sep. 2002, pp. 823–828.
\bibitem{c14}S. Hirose, T. Shirasu, and E. F. Fukushima, ``Proposal for cooperative robot ‘Gunryu’ composed of autonomous segments,'' in Robotics and Autonomous Systems, vol. 17, no. 1--2, P. Editor, Ed. New York: Elsevier, 1996, pp. 107--118.
\bibitem{c15}K. Osuka, T. Kinugasa, R. Hayashi, K. Yoshida, D. Owaki, and A. Ishiguro, ``Centipede type robot i-centipot: From machine to creatures,'' in Journal of Robotics and Mechatronics, vol. 31, no. 5, 2019, pp. 723–726.
\bibitem{c19} P. E. Hart, N. J. Nilsson, and B. Raphael, ``A formal basis for the heuristic determination of minimum cost paths,'' in IEEE Transactions on Systems Science and Cybernetics, vol. 4, no. 2, P. Editor, Ed. New York: IEEE, 1968, pp. 100--107.
\bibitem{c16}H. Yamada and S. Hirose, ``Development of practical 3-dimensional active cord mechanism ACM-R4,'' in Journal of Robotics and Mechatronics, vol. 18, no. 3, P. Editor, Ed. New York: Fuji Technology Press, 2006, pp. 305--311.
\bibitem{c17}S. Takaoka, H. Yamada, and S. Hirose, ``Snake-like active wheel robot ACM-R4.1 with joint torque sensor and limiter,'' in Intelligent Robots and Systems, 2011 IEEE/RSJ International Conference, P. Editor, Ed. New York: IEEE, 2011.
\bibitem{c18}L. E. Parker, et al., ``Tightly-coupled navigation assistance in heterogeneous multi-robot teams,'' in Intelligent Robots and Systems, 2004 IEEE/RSJ International Conference, vol. 1, P. Editor, Ed. New York: IEEE, 2004.

\end{thebibliography}
\end{document}